\documentclass[11pt,a4paper]{article}
\usepackage{naacl2021}
\usepackage{times}
\usepackage{amsmath}
\usepackage{amssymb}
\usepackage{amsthm}
\usepackage{booktabs}
\usepackage{latexsym}
\usepackage{stmaryrd}
\usepackage{xspace}
\usepackage{graphicx}

\newcommand{\std}[1]{\footnotesize\textcolor{gray}{#1}}

\usepackage{microtype}

\usepackage{xcolor}

\newcommand{\simplex}{\triangle}
\newcommand{\reals}{\mathbb{R}}

\newcommand{\langp}[2]{\textsc{#1}$\shortrightarrow$\textsc{#2}}
\newcommand{\langpb}[2]{\textsc{#1}$\leftrightarrow$\textsc{#2}}

\newcommand{\sts}{\emph{seq2seq}}
\newcommand{\bs}{\boldsymbol}
\newcommand*{\ie}{\textit{i.\hspace{.07em}e.}\@\xspace}

\DeclareMathOperator*{\argmax}{\mathsf{argmax}}
\DeclareMathOperator*{\softmax}{\mathsf{softmax}}
\DeclareMathOperator*{\acc}{\mathsf{acc}}
\DeclareMathOperator*{\conf}{\mathsf{conf}}

\usepackage{mdframed}
 \definecolor{theoremcolor}{rgb}{0.94, 0.94, 0.94}
\definecolor{examplecolor}{rgb}{1, 1, 1.0}
\mdfsetup{
    backgroundcolor=theoremcolor,
    linewidth=1pt,
}

\newmdtheoremenv{proposition}{Proposition}

\title{Smoothing and Shrinking the Sparse Seq2Seq Search Space}

\author{Ben Peters\textsuperscript{$\dag$} \and
        Andr\'e F.~T. Martins\textsuperscript{$\dag\ddag$} \\
\textsuperscript{$\dag$}Instituto de Telecomunica\c{c}\~oes, Lisbon, Portugal \\
\textsuperscript{$\ddag$}Unbabel, Lisbon, Portugal\\
\href{mailto:benzurdopeters@gmail.com}{\tt benzurdopeters@gmail.com},\quad
\href{mailto:andre.t.martins@tecnico.ulisboa.pt}{\tt andre.t.martins@tecnico.ulisboa.pt}
}

\date{}

\begin{document}
\maketitle
\begin{abstract}
Current sequence-to-sequence models are trained to minimize cross-entropy and use softmax to compute the locally normalized probabilities over target sequences.
While this setup has led to strong results in a variety of tasks, 
one unsatisfying aspect is its length bias: models give high scores to short, inadequate hypotheses and often make the empty string the argmax---the so-called \emph{cat got your tongue} problem.
Recently proposed entmax-based sparse sequence-to-sequence models present a possible solution, since they can shrink the search space by assigning zero probability to bad hypotheses, but their ability to handle word-level tasks with transformers has never been tested.
In this work, we show that entmax-based models effectively solve the \emph{cat got your tongue} problem, removing a major source of model error for neural machine translation. 
In addition, we generalize label smoothing, a critical regularization technique, to the broader family of Fenchel-Young losses, which includes both cross-entropy and the entmax losses.
Our resulting label-smoothed entmax loss models set a new state of the art on multilingual grapheme-to-phoneme conversion and deliver improvements and better calibration properties on cross-lingual morphological inflection and machine translation for 6 language pairs.
\end{abstract}

\section{Introduction}

Sequence-to-sequence models \citep[{\sts}:][]{sutskever2014sequence,bahdanau2014neural,vaswani2017attention} have become a powerful and flexible tool for a variety of NLP tasks, including machine translation (MT), morphological inflection \citep[MI; ][]{faruqui-etal-2016-morphological}, and grapheme-to-phoneme conversion \citep[G2P; ][]{yao2015sequence}.
These models often perform well, but they have a bias that favors short hypotheses.
This bias is problematic: it has been pointed out as the cause \citep{koehn-knowles-2017-six,yang-etal-2018-breaking,murray-chiang-2018-correcting} of the \emph{beam search curse}, in which increasing the width of beam search actually \emph{decreases} performance on neural machine translation (NMT).
Further illustrating the severity of the problem, \citet{stahlberg-byrne-2019-nmt} showed that the highest-scoring target sequence in NMT is often the empty string, a phenomenon they dubbed the \emph{cat got your tongue} problem. 
These results are undesirable because they show that NMT models' performance depends on the search errors induced by a narrow beam. 
It would be preferable for models to assign higher scores to good translations than to bad ones, rather than to depend on search errors to make up for model errors.

The most common way to alleviate this shortcoming is by altering the decoding objective
\citep{wu2016google,he2016improved,yang-etal-2018-breaking,meister-etal-2020-beam}, but this
does not address
the underlying problem: the model overestimates the probability of implausible hypotheses.
Other solutions use alternate training strategies \citep{murray-chiang-2018-correcting,shen-etal-2016-minimum}, but it would be preferable not to
change the training algorithm.

In this paper, 
we propose a solution based on sparse \emph{seq2seq} models \citep{peters-etal-2019-sparse}, which replace the output softmax \citep{bridle1990probabilistic} with the entmax transformation.
Entmax, unlike softmax, can learn locally sparse distributions over the target vocabulary.
This allows a sparse model to \textbf{shrink the search space}: 
that is, it can learn to give inadequate hypotheses zero probability, instead of counting on beam search to prune them.
This has already been demonstrated for MI, where the set of possible hypotheses
is often small enough to make beam search exact \citep{peters-etal-2019-sparse,peters-martins-2019-ist}.
We extend this analysis to MT: although exact beam search is not possible for this large vocabulary task, we show that entmax models prune many inadequate hypotheses, effectively solving the \emph{cat got your tongue} problem.

Despite this useful result, one drawback of entmax is that it is not compatible with label smoothing \citep{szegedy2016rethinking}, a useful regularization technique that is widely used for transformers \citep{vaswani2017attention}.
We solve this problem by generalizing label smoothing from the cross-entropy loss to the wider class of Fenchel-Young losses \citep{blondel2020learning}, which includes the entmax loss as a particular case.
We show that combining label smoothing with entmax loss improves results on both character- and word-level tasks while keeping the model sparse.
We note that, although label smoothing improves calibration, it also exacerbates the \emph{cat got your tongue} problem regardless of loss function.

To sum up, we make the following contributions:%
\footnote{Our code is available at \url{https://github.com/deep-spin/S7}.}

\begin{itemize}
    \item We show empirically that models trained with entmax loss rarely assign nonzero probability to the empty string, demonstrating that entmax loss is an elegant way to remove a major class of NMT model errors.
    \item We generalize label smoothing from  the cross-entropy loss to the wider class of Fenchel-Young losses, exhibiting a formulation for label smoothing which, to our knowledge, is novel.
    \item We show that Fenchel-Young label smoothing with entmax loss is highly effective on both character- and word-level tasks.
    Our technique allows us to set a new state of the art on the SIGMORPHON 2020 shared task for multilingual G2P \citep{Task1}.
    It also delivers improvements for crosslingual MI from SIGMORPHON 2019 \citep{mccarthy-etal-2019-sigmorphon} and for MT on IWSLT 2017 German $\leftrightarrow$ English \citep{cettolooverview}, KFTT Japanese $\leftrightarrow$ English \citep{neubig11kftt}, and WMT 2016 Romanian $\leftrightarrow$ English \citep{bojar-etal-2016-findings} compared to a baseline with unsmoothed labels.
\end{itemize}

\section{Background}
A \emph{seq2seq} model learns a probability distribution $p_{\bs{\theta}}(y\mid x)$ over sequences $y$ from a target vocabulary $V$, conditioned on a source sequence $x$.
This distribution is then used at decoding time to find the most likely sequence $\hat{y}$:
\begin{equation}\label{eqn:s2sinference}
\hat{y} = \argmax_{y \in V^*} p_{\bs{\theta}}(y \mid x),
\end{equation}
where $V^*$ is the Kleene closure of $V$. 
This is an intractable problem; \emph{seq2seq} models depend on heuristic search strategies, most commonly beam search \citep{reddy1977speech}.
Most \emph{seq2seq} models are locally normalized, with probabilities that decompose by the chain rule:
\begin{equation}\label{eqn:chainrule}
    p_{\bs{\theta}}(y \mid x) = \prod_{i=1}^{|y|} p_{\bs{\theta}}(y_i \mid x, y_{<i}).
\end{equation}
This factorization implies that the probability of a hypothesis being generated is monotonically nonincreasing in its length, which favors shorter sequences.
This phenomenon feeds the beam search curse because short hypotheses\footnote{We use ``hypothesis'' to mean any sequence that ends with the special end-of-sequence token.} are pruned from a narrow beam but survive a wider one.

The conditional distribution
$p_{\bs{\theta}}(y_i \mid x, y_{<i})$ is obtained by first computing a vector of scores (or ``logits'') $\bs{z} = \bs{f}_{\bs{\theta}}(x, y_{<i}) \in \reals^{|V|}$, where $\bs{f}_{\bs{\theta}}$ is parameterized by a neural network, and then 
applying  a transformation $\pi: \reals^{|V|} \rightarrow \simplex^{|V|}$, which maps scores to the probability simplex $\simplex^{|V|} := \{\bs{p} \in \reals^{|V|} \colon \bs{p} \geq 0, \|\bs{p}\|_1 = 1\}$. 
The usual choice for $\pi$ is softmax \citep{bridle1990probabilistic}, which returns strictly positive values, ensuring that all sequences $\in V^*$ have nonzero probability.
Coupled with the short sequence bias, this causes significant model error.

\paragraph{Sparse \emph{seq2seq} models.}
In a sparse model, the output softmax is replaced by a transformation $\pi$ from the entmax family \citep{peters-etal-2019-sparse}.
Like softmax, entmax transformations return a vector in the simplex and are differentiable (almost) everywhere.
However, unlike softmax, they are capable of producing \textbf{sparse probability distributions}.
Concretely, this is done by using the so-called ``$\beta$-exponential function'' \citep{Tsallis1988} in place of the exponential, where $\beta \ge 0$:
\begin{equation}
\exp_{\beta}(v) := \left\{
    \begin{array}{ll}
     	[1 + (1-\beta)v]_+^{1/(1-\beta)}, & \beta \ne 1 \\
        \exp(v), & \beta = 1.
    \end{array}
    \right.  
\end{equation}
The $\beta$-exponential function converges to the regular exponential when $\beta \rightarrow 1$. 
Entmax models assume that $p(y_i\mid x, y_{<i})$ results from an $\alpha$-entmax transformation of the scores $\bs{z}$, defined as
\begin{equation}\label{eq:tsallis_exp}
    [\alpha\text{-}\mathsf{entmax}(\bs{z})]_y := \exp_{2-\alpha}(z_y - \tau_\alpha(\bs{z})),
\end{equation}
where $\tau_\alpha(\bs{z})$ is a constant which ensures normalization. When $\alpha=1$, \eqref{eq:tsallis_exp} turns to a regular exponential function and $\tau_1(\bs{z}) = \log \sum_{y'=1}^{|V|} \exp(z_{y'})$ is the log-partition function, recovering softmax.
When $\alpha=2$, we recover sparsemax \citep{martins2016softmax}. 
For $\alpha \in \{1.5, 2\}$, fast algorithms to compute \eqref{eq:tsallis_exp} are available which are almost as fast as evaluating softmax. For other values of $\alpha$, slower bisection algorithms exist. 

Entmax transformations are sparse 
for any $\alpha > 1$, with higher values tending to produce sparser outputs.
This sparsity allows a model to assign exactly zero probability to implausible hypotheses.
For tasks where there is only one correct target sequence, this often allows the model to concentrate all probability mass into a small set of hypotheses, making search exact \citep{peters-martins-2019-ist}.
This is not possible for open-ended tasks like machine translation, but the model is still locally sparse, assigning zero probability to many hypotheses.
These hypotheses will never be selected \textbf{at any beam width}.

\paragraph{Fenchel-Young Losses.} 
Inspired by the softmax generalization above, \citet{blondel2020learning} provided a tool for \textbf{constructing} a convex loss function. 
Let $\Omega: \simplex^{|V|} \rightarrow \reals$ be a strictly convex regularizer which is symmetric, \textit{i.e.}, $\Omega(\Pi \bs{p}) = \Omega(\bs{p})$ for any permutation $\Pi$ and any $\bs{p} \in \simplex^{|V|}$.%
\footnote{It is instructive to think of $\Omega$ as a generalized negative entropy: for example, as shown in \citet[Prop. 4]{blondel2020learning}, strict convexity and symmetry imply that $\Omega$ is minimized by the uniform distribution. For a more comprehensive treatment of Fenchel-Young losses, see the cited work. 
} %
Equipped with $\Omega$, we can define a regularized prediction function $\hat{\pi}_\Omega : \reals^{|V|} \rightarrow \simplex^{|V|}$, with this form:
\begin{equation}\label{eqn:regpred}
    \hat{\pi}_\Omega(\bs{z}) = \argmax_{\bs{p} \in \simplex^{|V|}} \bs{z}^\top \bs{p} - \Omega(\bs{p})
\end{equation}
where $\bs{z} \in \reals^{|V|}$ is the vector of label scores (logits) and $\Omega: \simplex^{|V|} \rightarrow \reals$ is a regularizer.
Equation~\ref{eqn:regpred} recovers both softmax and entmax with particular choices of $\Omega$: the negative Shannon entropy, $\Omega(\bs{p})=\sum_{y\in {V}} p_y \log p_y$, recovers the variational form of softmax \citep{wainwright_2008}, while the negative Tsallis entropy \citep{Tsallis1988} with parameter $\alpha$, defined as 
\begin{equation}\label{eq:tsallis}
    \Omega_\alpha(\bs{p}) = \left\{
    \begin{array}{ll}
    \frac{1}{\alpha(\alpha-1)}\left(\sum_{y\in V} p_y^\alpha - 1\right), & \text{if $\alpha \ne 1$}\\
    \sum_{y \in V} p_y \log p_y, & \text{if $\alpha = 1$,}
    \end{array}
    \right.
\end{equation}
recovers the $\alpha$-entmax transformation in \eqref{eq:tsallis_exp}, as shown by \citet{peters-etal-2019-sparse}. 

Given the choice of $\Omega$, the Fenchel-Young loss function $L_\Omega$ is defined as
\begin{equation}\label{eqn:fyloss}
    L_\Omega(\bs{z}; \bs{q}) := \Omega^*(\bs{z}) + \Omega(\bs{q}) - \bs{z}^\top \bs{q},
\end{equation}
where $\bs{q}$ is a target distribution, most commonly a one-hot vector indicating the gold label, $\bs{q} = \bs{e}_{y^*} = [0, \ldots, 0, \underbrace{1}_{\text{$y^*$-th entry}}, 0, \ldots, 0]$, and 
$\Omega^*$ is the convex conjugate of $\Omega$, defined variationally as:
\begin{equation}
    \Omega^*(\bs{z}) := \max_{\bs{p} \in \simplex^{|V|}} \bs{z}^\top \bs{p} - \Omega(\bs{p}).
\end{equation}
The name stems from the Fenchel-Young inequality, which states that the quantity \eqref{eqn:fyloss} is non-negative \citep[Prop.~3.3.4]{borwein2010convex}. 
When $\Omega$ is the generalized negative entropy, 
the loss \eqref{eqn:fyloss} becomes the Kullback-Leibler divergence between $\bs{q}$ and $\softmax(\bs{z})$ \citep[KL divergence; ][]{kldiv}, which equals the cross-entropy when $\bs{q}$ is a one-hot vector. More generally, if $\Omega \equiv \Omega_\alpha$ is the negative Tsallis entropy \eqref{eq:tsallis}, we obtain the \textbf{$\alpha$-entmax loss} \citep{peters-etal-2019-sparse}. 

\begin{figure}
    \centering
    \includegraphics[width=0.9\columnwidth]{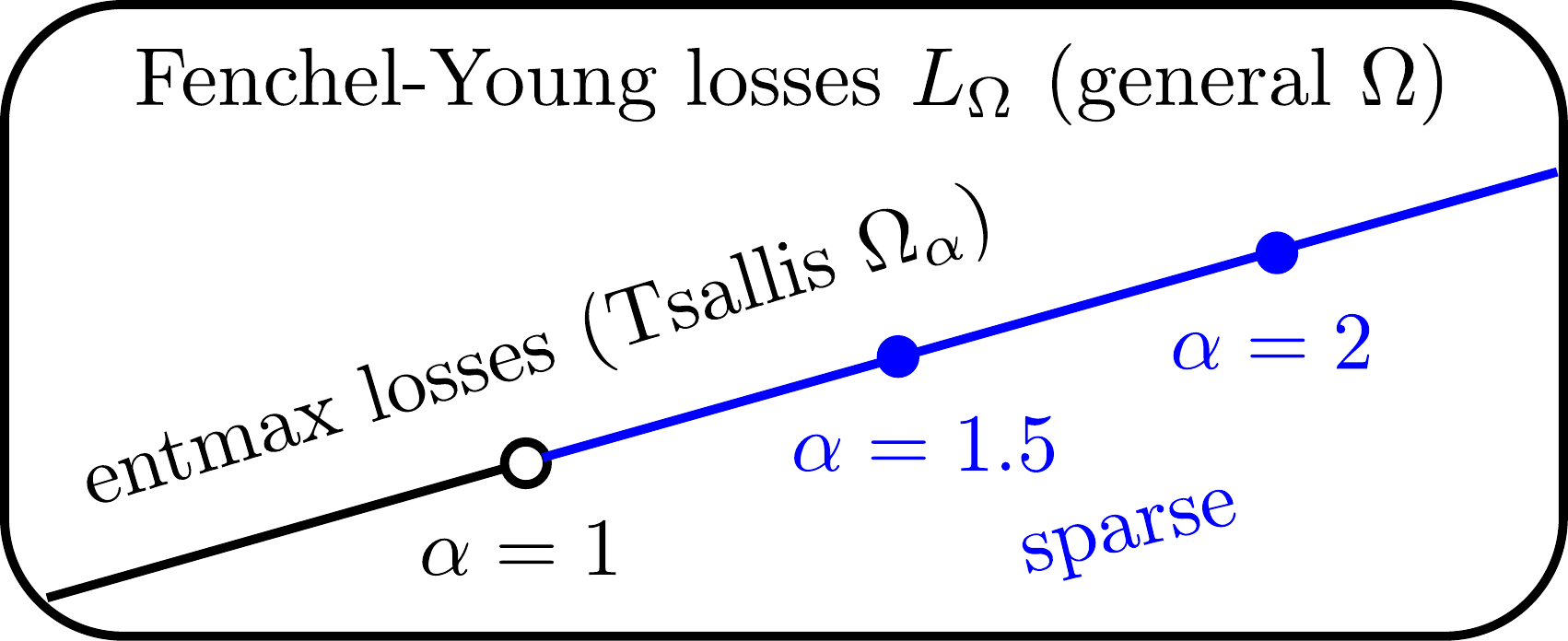}
    \caption{Diagram illustrating Fenchel-Young losses and the particular case of $\alpha$-entmax family. The case $\alpha=1$ corresponds to softmax and the cross-entropy loss, $\alpha=2$ to the sparsemax loss, and $\alpha=1.5$ to the 1.5-entmax loss. Any choice of $\alpha>1$ (in blue) can lead to sparse distributions.}
    \label{fig:fy_diagram}
\end{figure}

Fenchel-Young losses have nice properties for training neural networks with backpropagation: they are non-negative, convex, and differentiable as long as $\Omega$ is strictly convex \citep[Prop.~2]{blondel2020learning}. Their gradient is
\begin{equation}\label{eq:gradient_fy}
\nabla_{\bs{z}}  L_{\Omega}(\bs{z}; \bs{q}) = \hat{\pi}_\Omega(\bs{z}) - \bs{q},
\end{equation}
which generalizes the gradient of the cross-entropy loss. 
Figure~\ref{fig:fy_diagram} illustrates particular cases of Fenchel-Young losses considered in this paper.

\section{Fenchel-Young Label Smoothing}
Label smoothing \citep{szegedy2016rethinking} has become a popular technique for regularizing the output of a neural network.
The intuition behind it is that using the gold target labels from the training set can lead to overconfident models. 
To overcome this, label smoothing redistributes probability mass from the gold label to the other target labels.
When the redistribution is uniform, \citet{pereyra2017regularizing} and \citet{meister-etal-2020-generalized} pointed out that this is equivalent (up to scaling and adding a constant) to adding a second term to the loss that computes the KL divergence $D_\text{KL}(\bs{u} \| \bs{p}_{\bs{\theta}})$ between a uniform distribution $\bs{u}$ and the model distribution $\bs{p}_{\bs{\theta}}$.
While it might seem appealing to add a similar KL regularizer to a Fenchel-Young loss, this is not possible when $\bs{p}_{\bs{\theta}}$ contains zeroes because the KL divergence term becomes infinite.
This makes vanilla label smoothing \textbf{incompatible with sparse models}.
Fortunately, there is a more natural generalization of label smoothing to Fenchel-Young losses. For $\epsilon \in [0,1]$, we  define the {\bf Fenchel-Young label smoothing loss} as follows:
\begin{equation}\label{eq:fylabelsmoothing}
    L_{\Omega, \epsilon}(\bs{z}, \bs{e}_{y^*}) := L_{\Omega}(\bs{z}, (1-\epsilon)\bs{e}_{y^*}+\epsilon\bs{u}).
\end{equation}
The intuition is the same as in cross-entropy label smoothing: the target one-hot vector is mixed with a uniform distribution.

This simple definition leads to the following result, proved in Appendix~\ref{sec:proof_label_smoothing}:
\begin{proposition}\label{prop:label_smoothing}
\normalfont The Fenchel-Young label smoothing loss can be written as
\begin{equation}\label{eq:linreg}
L_{\Omega, \epsilon}(\bs{z}, \bs{e}_{y^*}) = L_\Omega(\bs{z}, \bs{e}_{y^*}) + \epsilon (z_{y^*} - \bar{z}) + C,
\end{equation}
where $C=-\Omega(\bs{e}_{y^*}) + \Omega((1-\epsilon)\bs{e}_{y^*}+\epsilon\bs{u})$ is a constant which does not depend on $\bs{z}$, and $\bar{z} := \bs{u}^\top \bs{z}$ is the average of the logits. 
Furthermore, up to a constant, we also have
\begin{equation}\label{eq:fyreg}
L_{\Omega, \epsilon}(\bs{z}, \bs{e}_{y^*}) \propto L_\Omega(\bs{z}, \bs{e}_{y^*}) + \lambda L_\Omega(\bs{z}, \bs{u}),
\end{equation}
where $\lambda = \frac{\epsilon}{1-\epsilon}$.
\end{proposition}

\begin{figure*}
    \centering
    \includegraphics[width=0.49\linewidth]{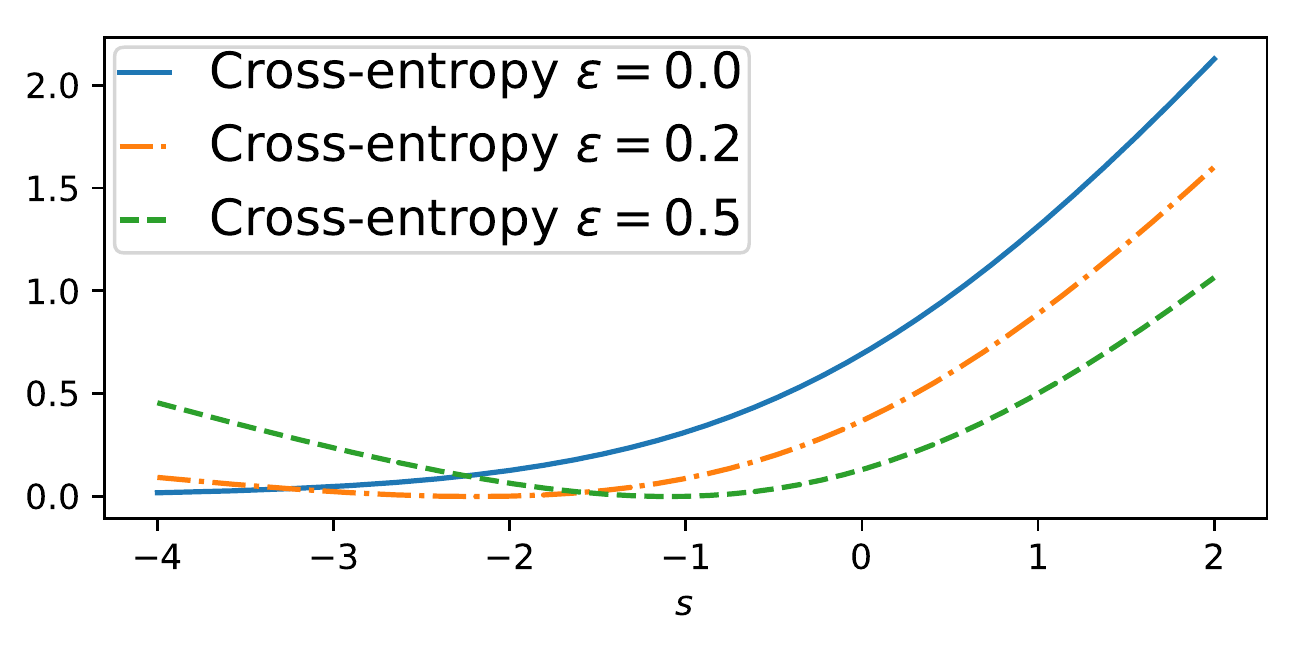}
    \includegraphics[width=0.49\linewidth]{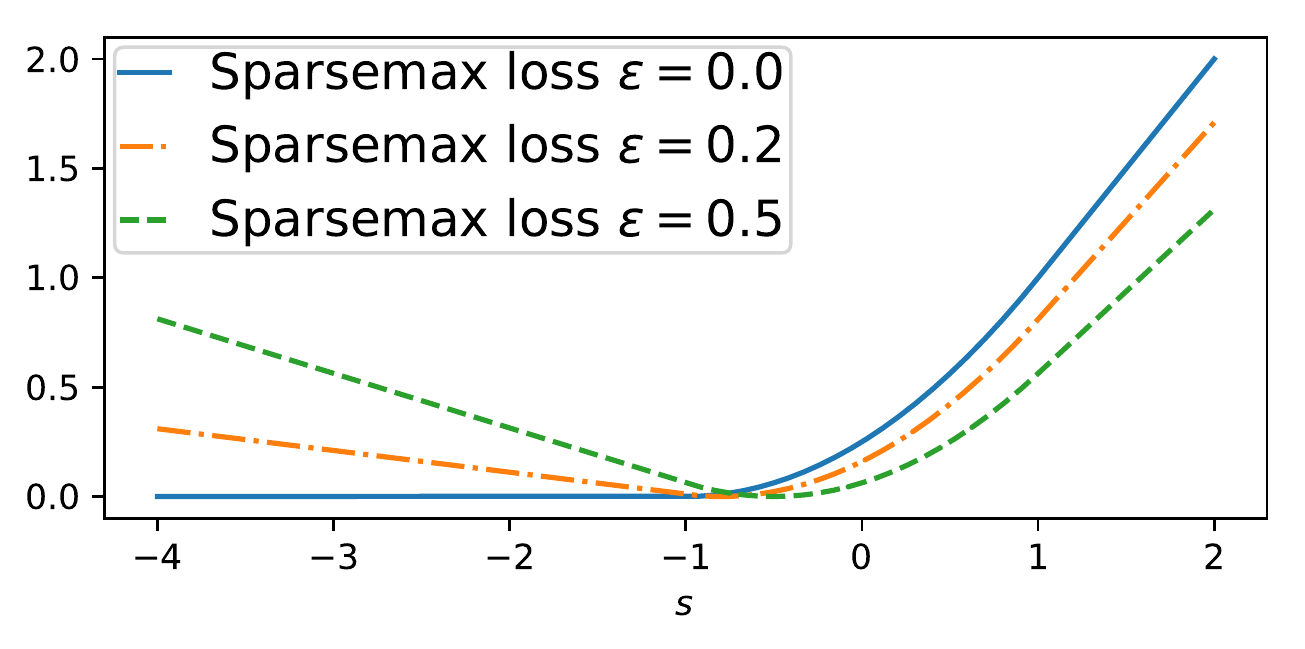}
    \caption{Fenchel-Young label smoothing with various $\epsilon$ values in the two-dimensional case for the cross-entropy loss (left) and sparsemax loss (right). In both cases we define $s = z_2 - z_1$ and assume $y^* = 1$.}
    \label{fig:smoothing-loss}
\end{figure*}

The first expression \eqref{eq:linreg} shows that, up to a constant, the smoothed Fenchel-Young loss equals the original loss plus a linear regularizer $\epsilon (z_{y^*} - \bar{z})$. While this regularizer can be positive or negative, we show in Appendix~\ref{sec:proof_label_smoothing} that its sum with the original loss $L_{\Omega}(\bs{z}, \bs{e}_{y^*})$ is always non-negative -- intuitively, if the score $z_{y^*}$ is below the average, resulting in negative regularization, the unregularized loss will also be larger, and the two terms balance each other. Figure~\ref{fig:smoothing-loss} shows the effect of this regularization in the graph of the loss -- we see that a correct prediction is linearly penalized with a slope of $\epsilon$; the larger the confidence, the larger the penalty.  
In particular, when $\Omega$ is the Shannon negentropy, this result shows a simple expression for vanilla label smoothing which, to the best of our knowledge, is novel.
The second expression \eqref{eq:fyreg} shows that it can also be seen as a form of regularization towards the uniform distribution. When $-\Omega$ is the Shannon entropy, the regularizer becomes a KL divergence and we obtain the interpretation of label smoothing for cross-entropy provided by \citet{pereyra2017regularizing} and \citet{meister-etal-2020-generalized}. Therefore, the same interpretation holds for the entire Fenchel-Young family if the regularization uses the corresponding Fenchel-Young loss with respect to a uniform.

\paragraph{Gradient of Fenchel-Young smoothed loss.} 

From Prop.~\ref{prop:label_smoothing} and Equation~\ref{eq:gradient_fy}, we immediately obtain the following expression for the gradient of the smoothed loss:
\begin{eqnarray}
\lefteqn{\nabla_{\bs{z}} L_{\Omega, \epsilon}(\bs{z}, \bs{e}_{y^*}) =}\nonumber\\ 
&=& \nabla_{\bs{z}} L_{\Omega}(\bs{z}, \bs{e}_{y^*}) + \epsilon(\bs{e}_{y^*} - \bs{u}) \nonumber\\
&=& \hat{\pi}_{\Omega}(\bs{z}) - (1-\epsilon)\bs{e}_{y^*} - \epsilon \bs{u},
\end{eqnarray}
that is, the computation of this gradient is straightforward by adding a constant vector to the original gradient of the Fenchel-Young loss; as the latter, it only requires the ability of computing the $\hat{\pi}_\Omega$ transformation, which is efficient in the entmax case as shown by \citet{peters-etal-2019-sparse}. Note that, unlike the gradient of the original entmax loss, the gradient of its smoothed version is not sparse (in the sense that it will not contain many zeroes); however, since $\bs{u}$ is the uniform distribution, it will contain many constant terms with value $-\epsilon/{|V|}$.

\section{Experiments}\label{sec:experiments}

We trained {\sts} models for three tasks: multilingual G2P, crosslingual MI, and MT.
These tasks present very different challenges.
In G2P and MI, character-level vocabularies are small
and there is usually only one correct target sequence.
The relative simplicity of these tasks is offset by the small quantity of training data and the strict evaluation: the model must produce exactly the right sequence.
This tests Fenchel-Young label smoothing's ability to learn \textit{exactly} in a low-resource setting.
On the other hand, MT is trained with much larger corpora and evaluated with less strict metrics, but uses subword vocabularies with sizes in the tens of thousands and has to manage more ambiguity because sentences typically have many correct translations.

In all tasks, we vary two hyperparameters:

\begin{itemize}
    \item Entmax Loss $\alpha$: this influences the sparsity of the probability distributions the model returns, with $\alpha=1$ recovering cross-entropy and larger $\alpha$ values encouraging sparser distributions. We use $\alpha \in \{1, 1.5, 2\}$ for G2P and MI, and $\alpha \in \{1, 1.5\}$ for MT.
    \item Fenchel-Young Label Smoothing $\epsilon$: higher values give more weight to the uniform smoothing distribution, discouraging sparsity.
    We use $\epsilon \in \{0, 0.01, 0.02, \dots, 0.15\}$ for G2P, $\epsilon \in \{0, 0.01, 0.05, 0.1\}$ for MI, and $\epsilon \in \{0, 0.01, 0.1\}$ for MT.
\end{itemize}

We trained all models with early stopping for a maximum of 30 epochs for MI and 100 epochs otherwise, keeping the best checkpoint according to a task-specific validation metric: Phoneme Error Rate for G2P, average Levenshtein distance for MI, and detokenized BLEU score for MT.
At test time, we decoded with a beam width of 5.
Our PyTorch code \citep{pytorch} is based on JoeyNMT \citep{JoeyNMT} and the entmax implementation from the entmax package.\footnote{\url{https://github.com/deep-spin/entmax}}

\subsection{Multilingual G2P}

\begin{table*}[t]

\begin{center}
\begin{tabular}{llrrrr}
\toprule
 & & \multicolumn{2}{c}{Single} & \multicolumn{2}{c}{Ensemble} \\
$\alpha$ & $\epsilon$ & WER $\downarrow$ & PER $\downarrow$ & WER $\downarrow$ & PER $\downarrow$ \\
\midrule
1 & 0 & 18.14 \std{$\pm$ 2.87} & 3.95 \std{$\pm$ 1.24} & 14.74 & 2.96 \\
  & 0.15 & 15.55 \std{$\pm$ 0.48} & 3.09 \std{$\pm$ 0.10} & 13.87 & 2.77 \\
\midrule
1.5 & 0 & 15.25 \std{$\pm$ 0.25} & 3.05 \std{$\pm$ 0.03} & 13.79 & 2.77 \\
 & 0.04 & 14.18 \std{$\pm$ 0.24} & 2.86 \std{$\pm$ 0.05} & \textbf{13.47} & \textbf{2.69} \\
\midrule
2 & 0 & 15.08 \std{$\pm$ 0.28} & 3.04 \std{$\pm$ 0.06} & 13.84 & 2.75 \\
 & 0.04 & 14.17 \std{$\pm$ 0.20} & 2.88 \std{$\pm$ 0.04} & 13.51 & 2.73 \\
\midrule
\multicolumn{2}{l}{\citep{yu-etal-2020-ensemble}} & & & 13.81 & 2.76 \\
\bottomrule
\end{tabular}

\end{center}
\caption{Multilingual G2P results on the SIGMORPHON 2020 Task 1 test set, macro-averaged across languages. Numbers in the Single column are the average of five independent model runs. The same models are used for the ensembles. Note that $\alpha=1$ and $\alpha=2$ recover cross-entropy and sparsemax loss, respectively.\label{table:G2P}}
\end{table*}

\paragraph{Data.}
We use the data from SIGMORPHON 2020 Task 1~\citep{Task1}, which includes 3600 training examples in each of 15 languages.
We train a single multilingual model \citep[following][]{peters-martins-2020-one}
which must learn to apply spelling rules from several writing systems.

\paragraph{Training.}
Our models are similar to \citet{peters-martins-2020-one}'s RNNs, but with entmax 1.5 attention,
and language embeddings only in the source. 

\paragraph{Results.}
Multilingual G2P results are shown in Table~\ref{table:G2P}, along with the best previous result \citep{yu-etal-2020-ensemble}.
We report two error metrics, each of which is computed per-language and averaged:
\begin{itemize}
    \item Word Error Rate (WER) is the percentage of hypotheses which do not exactly match the reference. 
    This harsh metric gives no credit for partial matches.
    \item Phoneme Error Rate (PER) is the sum of Levenshtein distances between each hypothesis and the corresponding reference, divided by the total length of the references.
\end{itemize}
These results show that the benefits of sparse losses and label smoothing can be combined.
Individually, both label smoothing and sparse loss functions ($\alpha > 1$) consistently improve over unsmoothed cross-entropy ($\alpha = 1$).
Together, they produce the best reported result on this dataset.
Our approach is very simple, as it requires manipulating only the loss function: there are no changes to the standard {\sts} training or decoding algorithms, no language-specific training or tuning, and no external auxiliary data.
In contrast, the previous state of the art \citep{yu-etal-2020-ensemble} relies on a complex self-training procedure in which a genetic algorithm is used to learn to ensemble several base models.

\subsection{Crosslingual MI}

\begin{table}[t]

\begin{center}
\begin{tabular}{llrr}
\toprule
$\alpha$ & $\epsilon$ & Acc. $\uparrow$ & Lev. Dist. $\downarrow$ \\
\midrule
1 & 0 & 50.16 & 1.12\\
 & $>0$ & 52.72 & 1.04\\
\midrule
1.5 & 0 & 55.21 & 0.98\\
 & $>0$ & 57.40 & 0.92\\
\midrule
2 & 0 & 56.01 & 0.97\\
 & $>0$ & 57.77 & \textbf{0.90}\\
\midrule
\multicolumn{2}{l}{CMU-03} & \textbf{58.79} & 1.52\\
\bottomrule
\end{tabular}

\end{center}
\caption{Macro-averaged MI results on the SIGMORPHON 2019 Task 1 test set. When $\epsilon > 0$, it is tuned separately for each language pair. 
\label{table:morph}}
\end{table}

\paragraph{Data.}
Our data come from SIGMORPHON 2019 Task 1 \citep{mccarthy-etal-2019-sigmorphon}, which includes datasets for 100 language pairs.
Each training set combines roughly 10,000 examples from a high resource language with 100 examples from a (simulated) low resource language.\footnote{Although most of the low resource sets are for languages that lack real-world NLP resources, others are simply small training sets in widely-spoken languages such as Russian.}
Development and test sets only cover the low resource language. 

\paragraph{Training.}
We reimplemented \textsc{GatedAttn} \citep{peters-martins-2019-ist}, an RNN model with separate encoders for lemma and morphological tags.
We copied their hyperparameters, except that we used two layers for all encoders.
We concatenated the high and low resource training data.
In order to make sure the model paid attention to the low resource training data, we either oversampled it 100 times or used data hallucination \citep{anastasopoulos-neubig-2019-pushing} to generate synthetic examples.
Hallucination worked well for some languages but not others, so we treated it as a hyperparameter.

\paragraph{Results.}
We compare to CMU-03\footnote{We specifically use the official task numbers from \citet{mccarthy-etal-2019-sigmorphon}, which are more complete than those reported in \citet{anastasopoulos-neubig-2019-pushing}.} \citep{anastasopoulos-neubig-2019-pushing}, a two-encoder model with a sophisticated multi-stage training schedule.
Despite our models' simpler training technique, they performed nearly as well in terms of accuracy, while recording, to our knowledge, the best Levenshtein distance on this dataset.

\subsection{Machine Translation}

\begin{table*}[t]

\begin{center}
\begin{tabular}{llrrrrrrr}
\toprule
$\alpha$ & $\epsilon$& \langp{de}{en} & \langp{en}{de} & \langp{ja}{en} & \langp{en}{ja} & \langp{ro}{en} & \langp{en}{ro}\\
\midrule
1 & 0 & 27.05 \std{$\pm$ 0.05} & 23.36 \std{$\pm$ 0.10} & 20.52 \std{$\pm$ 0.13} & 26.94 \std{$\pm$ 0.32} & 29.41 \std{$\pm$ 0.20} & 22.84 \std{$\pm$ 0.12}\\
& $>0$ & 27.72 \std{$\pm$ 0.11} & 24.24 \std{$\pm$ 0.28} & 20.99 \std{$\pm$ 0.12}& 27.28 \std{$\pm$ 0.17} & 30.03 \std{$\pm$ 0.05} & 23.15 \std{$\pm$ 0.27}\\
1.5 & 0 & \textbf{28.12} \std{$\pm$ 0.01} & 24.03 \std{$\pm$ 0.06} & 21.23 \std{$\pm$ 0.10} & \textbf{27.58} \std{$\pm$ 0.34} & 30.27 \std{$\pm$ 0.16} & \textbf{23.74} \std{$\pm$ 0.08}\\
& $>0$ & 28.11 \std{$\pm$ 0.16} & \textbf{24.36} \std{$\pm$ 0.12} & \textbf{21.34} \std{$\pm$ 0.08} & \textbf{27.58} \std{$\pm$ 0.16} & \textbf{30.37} \std{$\pm$ 0.04} & 23.47 \std{$\pm$ 0.04}\\
\bottomrule
\end{tabular}

\end{center}
\caption{MT results, averaged over three runs. For label smoothing, we select  the best $\epsilon$ on the development set.\label{table:mt} }
\end{table*}

Having shown the effectiveness of our technique on character-level tasks, we next turn to MT.
To our knowledge, entmax loss has never been used for transformer-based MT;
\citet{correia-etal-2019-adaptively} used entmax only for transformer attention.

\paragraph{Data.}
We used joint BPE \citep{sennrich-etal-2016-neural} with 32,000 merges for these language pairs:
\begin{itemize}
    \item IWSLT 2017 German $\leftrightarrow$ English \citep[\langpb{de}{en}, ][]{cettolooverview}: 200k training examples.
    \item KFTT Japanese $\leftrightarrow$ English \citep[\langpb{ja}{en},][]{neubig11kftt}: 330k training examples.
    \item WMT 2016 Romanian $\leftrightarrow$ English \citep[\langpb{ro}{en},][]{bojar-etal-2016-findings}: 610k training examples.
\end{itemize}

\paragraph{Training.}
We trained transformers with the base
dimension and layer settings
\citep{vaswani2017attention}.
We optimized with Adam \citep{kingma2014adam} and used Noam scheduling with 10,000 warmup steps.
The batch size was 8192 tokens.

\paragraph{Results.}
Table~\ref{table:mt} reports our models' performance in terms of untokenized BLEU \citep{papineni-etal-2002-bleu}, which we computed with SacreBLEU \citep{post2018call}.
The results show a clear advantage for label smoothing and entmax loss, both separately and together: 
label-smoothed entmax loss is the best-performing configuration on 3 out of 6 language pairs, unsmoothed entmax loss performs best on 2 out of 6, and they tie on the remaining one.
Although label-smoothed cross-entropy is seen as an essential ingredient for transformer training, \textbf{entmax loss models beat it even without label smoothing} for every pair except \langp{en}{de}. 

\section{Analysis}

\begin{table*}[t]

\begin{center}
\begin{tabular}{llrrrrrr}
\toprule
$\alpha$ & $\epsilon$ & \langp{de}{en} & \langp{en}{de} & \langp{ja}{en} & \langp{en}{ja} & \langp{ro}{en} & \langp{en}{ro}\\
\midrule
1 & 0 & 8.07 \std{$\pm$ 1.21} & 12.97 \std{$\pm$ 2.58} & 23.10 \std{$\pm$ 1.01} & 14.38 \std{$\pm$ 1.06} & 9.10 \std{$\pm$ 0.82} & 4.32 \std{$\pm$ 0.20}\\
 & 0.01 & 9.66 \std{$\pm$ 1.55} & 17.87 \std{$\pm$ 1.63} & 22.96 \std{$\pm$ 1.28} & 15.41 \std{$\pm$ 1.62} & 12.19 \std{$\pm$ 0.98} & 17.56 \std{$\pm$ 5.55}\\
 & 0.1 & 22.17 \std{$\pm$ 1.79} & 30.25 \std{$\pm$ 1.54} & 34.79 \std{$\pm$ 1.12} & 31.19 \std{$\pm$ 1.80} & 54.01 \std{$\pm$ 14.54} & 47.24 \std{ $\pm$ 14.66}\\
1.5 & 0 & 0.50 \std{$\pm$ 0.03} & 0.78 \std{$\pm$ 0.71} & 0.03 \std{$\pm$ 0.04} & 0.63 \std{$\pm$ 0.66} & 0.18 \std{$\pm$ 0.12} & 0.88 \std{$\pm$ 0.46}\\
& 0.01 & 1.11 \std{$\pm$ 0.70} & 6.27 \std{$\pm$ 5.37} & 2.00 \std{$\pm$ 2.59} & 1.57 \std{$\pm$ 1.74} & 2.70 \std{$\pm$ 2.26} & 0.92 \std{$\pm$ 0.44}\\
& 0.1 & 0.96 \std{$\pm$ 0.55} & 0.65 \std{$\pm$ 0.30} & 15.67 \std{$\pm$ 22.16} & 13.89 \std{$\pm$ 19.53} & 35.13 \std{$\pm$ 24.33} & 44.12 \std{$\pm$ 2.39}\\
\bottomrule
\end{tabular}

\end{center}
\caption{Percentage of development set examples for which the model assigns higher probability to the empty string than to the beam-decoded hypothesis.
\label{table:emptyvsbeam} }
\end{table*}

\paragraph{Model error.}
\citet{stahlberg-byrne-2019-nmt} showed
that the bias in favor of short strings is so strong for softmax NMT models that the argmax
sequence is usually the empty string.
However,
they did not consider the impact
of sparsity or label smoothing.\footnote{They trained with ``transformer-base'' settings, implying label smoothing, and did not compare to unsmoothed losses.}
We show in Table~\ref{table:emptyvsbeam} how often the empty string is more probable than the beam search hypothesis.
This is an \textbf{upper bound} for how often the empty string is the argmax because there can also be other hypotheses that are more probable than the empty string.
The results show that $\alpha$ and $\epsilon$ both matter: sparsity substantially reduces the frequency with which the empty string is more probable than the beam search hypothesis,
while label smoothing usually increases it.
Outcomes vary widely
with $\alpha=1.5$ and $\epsilon=0.1$: \langpb{de}{en} models did not seriously suffer from the problem, \langp{en}{ro} did,
and the other three language pairs
differed
from one run to another.
The optimal label smoothing value with cross-entropy is invariably $\epsilon=0.1$, which encourages the \emph{cat got your tongue} problem; on the other hand, entmax loss does better with $\epsilon=0.01$ for every pair except \langp{ro}{en}. 

\paragraph{Label smoothing and sparsity.}
\begin{table*}[t]
    \centering
    \begin{tabular}{lrrrrrr}
\toprule
$\epsilon$ &  \langp{de}{en} &  \langp{en}{de} &  \langp{ja}{en} &  \langp{en}{ja} &  \langp{ro}{en} &  \langp{en}{ro} \\
\midrule
0 &           0.09 &           0.08 &           0.12 &           0.08 &           0.11 &           0.07 \\
0.01 &           0.17 &           0.14 &           0.19 &           0.15 &           0.21 &           0.18 \\
0.1 &           8.74 &           5.67 &           6.13 &           7.62 &           1.98 &           0.84 \\
\bottomrule
\end{tabular}
    \caption{Average percentage of the target vocabulary with nonzero probability with forced decoding for entmax loss MT models. For cross-entropy models, this is always 100\%.}
    \label{table:density}
\end{table*}

\citet{peters-etal-2019-sparse} previously showed that RNN models trained with entmax loss become locally very sparse.
Table~\ref{table:density} shows that this is
true of
transformers as well.
Label smoothing encourages greater density,
although for the densest language pair (\langp{de}{en}) this only equates to an average support size of roughly 1500 out of a vocabulary of almost 18,000 word types.
The relationship between density and overestimating the empty string is inconsistent with $\epsilon=0.1$: \langpb{de}{en} models 
become much more dense but
rarely
overestimate the empty string (Table~\ref{table:emptyvsbeam}).
The opposite occurs for \langpb{ro}{en}: models with $\epsilon=0.1$ become only slightly more dense but are much more prone to model error.
This suggests that corpus-specific factors
influence
both sparsity and how easily
bad hypotheses can be pruned.

\paragraph{Calibration.}
\begin{table*}[t]
    \centering
\begin{tabular}{llrrrrrr}
\toprule
  $\alpha$       &   $\epsilon$   &  \langp{de}{en} &  \langp{en}{de} &  \langp{ja}{en} &  \langp{en}{ja} &  \langp{ro}{en} &  \langp{en}{ro} \\
\midrule
1 & 0 &     0.146 &     0.149 &     0.186 &     0.167 &     0.166 &     0.188 \\
         & 0.01 &     0.147 &     0.131 &     0.175 &     0.162 &     0.160 &     0.176 \\
         & 0.1 &     0.078 &     0.077 &     0.116 &     0.095 &     0.102 &     0.125 \\
1.5 & 0 &     0.123 &     0.110 &     0.147 &     0.147 &     0.133 &     0.152 \\
         & 0.01 &     0.113 &     0.090 &     0.145 &     0.141 &     0.132 &     0.151 \\
         & 0.1 &     0.049 &     0.039 &     0.099 &     0.098 &     0.102 &     0.123 \\
\bottomrule
\end{tabular}
    \caption{MT development set Expected Calibration Error with 10 bins. Lower values indicate better calibration.}
    \label{table:calibration}
\end{table*}
This is the degree to which a model's confidence about its predictions (\ie class probabilities)
accurately measure
how likely those predictions are to be correct.
It has been shown \citep{muller2019does,kumar2019calibration}
that label smoothing improves the calibration of {\sts} models.
We computed
the Expected Calibration Error \citep[ECE; ][]{naeini2015obtaining}\footnote{
$ECE = \sum_{m=1}^M \frac{|B_m|}{N} |\acc(B_m) - \conf(B_m)|$ partitions the model's $N$ force-decoded predictions into $M$ evenly-spaced bins and computes the difference between the accuracy ($\acc(B_m)$) and the average probability of the most likely prediction ($\conf(B_m)$) within that bin. We use $M=10$.} of our MT models and confirmed their findings.
Our results, in Table~\ref{table:calibration}, also show that
\textbf{sparse models are better calibrated} than their dense counterparts.
This shows that entmax models do not become overconfident even though probability mass is usually concentrated in a small
set
of possibilities.
The good calibration of label smoothing may seem surprising in light of Table~\ref{table:emptyvsbeam}, which shows that label-smoothed models
overestimate
the probability of
inadequate hypotheses.
However, ECE depends only on the relationship between
model
accuracy and the
score
of the \textbf{most likely} label.
This
shows
the
tradeoff: larger $\epsilon$ values limit overconfidence
but make
the tail heavier.
Setting $\alpha = 1.5$ with a moderate $\epsilon$ value seems to be a sensible
balance.

\section{Related Work}
\paragraph{Label smoothing.} Our work fits into a larger family of techniques that penalize model overconfidence.
\citet{pereyra2017regularizing} proposed the confidence penalty, which reverses the direction of the KL divergence in the smoothing expression.
\citet{meister-etal-2020-generalized} then introduced a parameterized family of generalized smoothing techniques, different from Fenchel-Young Label Smoothing, which recovers vanilla label smoothing and the confidence penalty as special cases.
In a different direction, \citet{wang-etal-2020-inference} improved inference calibration with a graduated label smoothing technique that uses 
larger smoothing weights for predictions that a baseline model is more confident of.
Other works have smoothed over sequences instead of tokens \citep{norouzi2016reward,elbayad-etal-2018-token,lukasik-etal-2020-semantic}, but this requires approximate techniques
for deciding which sequences to smooth.

\paragraph{MAP decoding and the empty string.}
We showed that sparse distributions suffer less from the \emph{cat got your tongue} problem than their dense counterparts.
This makes sense in light of the finding that exact MAP decoding works for MI, where probabilities are very peaked even with softmax \citep{forster-meister-2020-sigmorphon}.
For
tasks like MT, this is not the case: \citet{eikema2020map} pointed out that the
argmax
receives so little mass that it is almost arbitrary, so seeking it with MAP decoding
(which beam search approximates)
itself causes many deficiencies of decoding.
On the other hand, \citet{meister-etal-2020-beam} showed that
beam search has a helpful bias
and introduced regularization penalties for MAP decoding that encode
it
explicitly.
Entmax neither directly addresses the faults of MAP decoding nor
compensates
for the locality biases of beam search, instead
shrinking
the gap between beam search and exact decoding.
It
would be interesting,
however, to experiment with these two approaches with entmax in place of softmax. 

\section{Conclusion}
We generalized label smoothing from cross-entropy to the wider class of Fenchel-Young losses. When combined with the entmax loss, we showed meaningful gains on character and word-level tasks, including a new state of the art on multilingual G2P. In addition, we showed that the ability of entmax to shrink the search space significantly alleviates the \emph{cat got your tongue} problem in machine translation, while also improving model calibration.

\bibliography{anthology,naacl2021}
\bibliographystyle{acl_natbib}

\appendix

\clearpage

\onecolumn





\section{Proof of Proposition~\ref{prop:label_smoothing}}\label{sec:proof_label_smoothing}

For full generality, we consider label smoothing with an arbitrary distribution $\bs{r} \in \simplex^{|V|}$, which may or not be the uniform distribution. We also consider an arbitrary gold distribution $\bs{q} \in \simplex^{|V|}$, not necessarily a one-hot vector. Later we will particularize for the case $\bs{r} = \bs{u} = [1/{|V|}, \ldots, 1/{|V|}]$ and $\bs{q} = \bs{e}_{y^*}$, the case of interest in this paper.

For this general case, the Fenchel-Young label smoothing loss is defined analogously to \eqref{eq:fylabelsmoothing} as
\begin{eqnarray}
L_{\Omega, \epsilon, \bs{r}}(\bs{z}; \bs{q}) := L_{\Omega}(\bs{z}, (1-\epsilon)\bs{q} + \epsilon \bs{r}).
\end{eqnarray}

A key quantity in this proof is the \textbf{Bregman information} induced by $\Omega$ (a generalization of the Jensen-Shannon divergence, see \citet[\S 3.1]{banerjee2005clustering} and \citet[\S 3.3]{blondel2020learning}): 
\begin{equation}\label{eq:bregman_information}
I_{\Omega, \epsilon}(\bs{q}, \bs{r}) := -\Omega((1-\epsilon)\bs{q} + \epsilon \bs{r}) + (1-\epsilon)\Omega(\bs{q}) + \epsilon \Omega(\bs{r}).
\end{equation}
Note that, from the convexity of $\Omega$ and Jensen's inequality, we always have $I_{\Omega, \epsilon}(\bs{q}; \bs{r}) \ge 0$. 

Using the definition of Fenchel-Young loss \eqref{eqn:fyloss}, we obtain
\begin{eqnarray}
L_{\Omega, \epsilon, \bs{r}}(\bs{z}; \bs{q}) 
&=& \Omega^*(\bs{z}) - \bs{z}^\top((1-\epsilon)\bs{q} + \epsilon \bs{r}) + \Omega((1-\epsilon)\bs{q} + \epsilon \bs{r}) \nonumber\\
&=& (1-\epsilon)\underbrace{\left(\Omega^*(\bs{z}) - \bs{z}^\top \bs{q} + \Omega(\bs{q})\right)}_{L_\Omega(\bs{z}; \bs{q})}
+ \epsilon \underbrace{\left(\Omega^*(\bs{z}) - \bs{z}^\top \bs{r} + \Omega(\bs{r})\right)}_{L_\Omega(\bs{z}; \bs{r})} 
+ \nonumber\\ &&
\underbrace{\Omega((1-\epsilon)\bs{q} + \epsilon \bs{r})
- (1-\epsilon)\Omega(\bs{q}) - \epsilon \Omega(\bs{r})}_{:= - I_{\Omega, \epsilon}(\bs{q}; \bs{r})}\nonumber\\
&=& (1-\epsilon)L_\Omega(\bs{z}; \bs{q}) + \epsilon L_\Omega(\bs{z}; \bs{r}) - I_{\Omega, \epsilon}(\bs{q}; \bs{r}).
\end{eqnarray}
Therefore, up to a constant term (with respect to $\bs{z}$) and a scalar multiplication by $1-\epsilon$, the Fenchel-Young smoothing loss is nothing but the original Fenchel-Young loss regularized by $L_\Omega(\bs{z}; \bs{r})$, with regularization constant $\lambda = \frac{\epsilon}{1-\epsilon}$, as stated in Proposition~\ref{prop:label_smoothing}. 
This generalizes the results of \citet{pereyra2017regularizing} and \citet{meister-etal-2020-generalized}, obtained as a particular case when $\Omega$ is the negative Shannon entropy and $L_\Omega$ is the Kullback-Leibler divergence.

We now derive the expression \eqref{eq:linreg}: 
\begin{eqnarray}\label{eq:smoothed_fyloss_2}
L_{\Omega, \epsilon, \bs{r}}(\bs{z}, \bs{q}) &=&  (1-\epsilon)L_\Omega(\bs{z}; \bs{q}) + \epsilon L_\Omega(\bs{z}; \bs{r}) - I_{\Omega, \epsilon}(\bs{q}; \bs{r})\nonumber\\
&=&  (1-\epsilon)L_\Omega(\bs{z}; \bs{q}) + \epsilon \underbrace{\left( \Omega^*(\bs{z}) - \bs{z}^\top \bs{r} + \Omega(\bs{r}) \right)}_{L_\Omega(\bs{z}; \bs{q}) + \bs{z}^\top \bs{q} -  \Omega(\bs{q}) - \bs{z}^\top \bs{r} + \Omega(\bs{r})} - I_{\Omega, \epsilon}(\bs{q}; \bs{r})\nonumber\\
&=& L_\Omega(\bs{z}; \bs{q}) + \epsilon (\bs{z}^\top \bs{q}  - \bs{z}^\top \bs{r}) + \epsilon(\Omega(\bs{r}) - \Omega(\bs{q})) - I_{\Omega, \epsilon}(\bs{q}; \bs{r})\nonumber\\
&=& L_\Omega(\bs{z}; \bs{q}) + \epsilon (\bs{z}^\top \bs{q}  - \bs{z}^\top \bs{r}) \underbrace{-\Omega(\bs{q}) + \Omega((1-\epsilon)\bs{q} + \epsilon \bs{r})}_{:= C \text{\quad (constant)}}.
\end{eqnarray} 
If $-\Omega(\bs{q}) \le -\Omega(\bs{r})$ (\textit{i.e.}, if the regularizing distribution $\bs{r}$ has higher generalized entropy than the model distribution $\bs{q}$, as is expected from a regularizer), then
\begin{eqnarray}
C &=& -\Omega(\bs{q}) + \Omega((1-\epsilon)\bs{q} + \epsilon \bs{r}) \nonumber\\
&=& -(1-\epsilon)\Omega(\bs{q}) - \epsilon \Omega(\bs{q}) + \Omega((1-\epsilon)\bs{q} + \epsilon \bs{r}) \nonumber\\
&\le & -(1-\epsilon)\Omega(\bs{q}) - \epsilon \Omega(\bs{r}) + \Omega((1-\epsilon)\bs{q} + \epsilon \bs{r}) \nonumber\\
&= & -I_{\Omega, \epsilon}(\bs{q}, \bs{r})
\nonumber\\
&\le & 0.
\end{eqnarray}
Since the left hand side of \eqref{eq:smoothed_fyloss_2} is by definition a Fenchel-Young loss, it must be non-negative. This implies that
\begin{equation}
    L_\Omega(\bs{z}; \bs{q}) + \epsilon (\bs{z}^\top \bs{q}  - \bs{z}^\top \bs{r}) \ge 0.
\end{equation}
In the conditions of the paper, we have $\bs{q} = \bs{e}_{y^*}$ and $\bs{r} = \bs{u}$, which satisfies the condition $-\Omega(\bs{q}) \le -\Omega(\bs{r})$ (this is implied by \citet[Prop. 4]{blondel2020learning} and the fact that $\Omega$ is strictly convex and symmetric). 
In this case, $\bs{z}^\top \bs{q} = z_{y^*}$ is the score of the gold label and $\bs{z}^\top \bs{r} = \frac{1}{|V|}\bs{z} = \bar{z}$ is the average score.

\end{document}